\documentclass[10pt,twocolumn,letterpaper]{article}

\usepackage{cvpr}              % To produce the CAMERA-READY version
\usepackage{cuted}
\usepackage{float}
\definecolor{cvprblue}{rgb}{0.21,0.49,0.74}
\usepackage[pagebackref,breaklinks,colorlinks,allcolors=cvprblue]{hyperref}
\usepackage{bm} 
\usepackage{multirow}
\usepackage{xcolor}
\usepackage{hyperref}       % hyperlinks
\usepackage{url}            % simple URL typesetting
\usepackage{booktabs}       % professional-quality tables
\usepackage{multirow}
\usepackage{amsfonts}       % blackboard math symbols
\usepackage{nicefrac}       % compact symbols for 1/2, etc.
\usepackage{microtype}      % microtypography
\usepackage{xcolor}         % colors
\usepackage{graphicx} 
\usepackage{amsmath} 
\usepackage{subcaption}
\usepackage{algorithm}
\usepackage{algorithmicx}
\usepackage{algpseudocode}
\usepackage{bm} 
\definecolor{third_color}{rgb}{0.3, 1, 0.7}

\usepackage[utf8]{inputenc}
\usepackage{textcomp} % Might help with some symbols
\usepackage{newunicodechar}

\newunicodechar{↔}{\(\leftrightarrow\)}
\def\paperID{4621} % *** Enter the Paper ID here
\def\confName{CVPR}
\def\confYear{2025}

\title{Uni-Renderer: Unifying Rendering and Inverse Rendering Via Dual Stream Diffusion}

\author{Zhifei Chen\textsuperscript{\rm 1,$*$}, 
Tianshuo Xu\textsuperscript{\rm 1,$*$}, 
Wenhang Ge\textsuperscript{\rm 1, $*$}, 
Leyi Wu\textsuperscript{\rm 1}, 
Dongyu Yan\textsuperscript{\rm 1}, 
Jing He\textsuperscript{\rm 1}, 
Luozhou Wang\textsuperscript{\rm 1},\\
Lu Zeng\textsuperscript{\rm 3}, 
Shunsi Zhang\textsuperscript{\rm 3}, 
Yingcong Chen\textsuperscript{\rm 1,2,$\dag$},\\
\textsuperscript{\rm 1}HKUST(GZ), \textsuperscript{\rm 2}HKUST, \textsuperscript{\rm 3}Quwan\\
{\tt\small \{zchen379, txu647\}@connect.hkust-gz.edu.cn; yingcongchen@ust.hk}
}

\begin{document}
\maketitle

% \begin{figure*}[tbph]
% \begin{center}
% \includegraphics[width=1\linewidth]{Figs/teaser.pdf}
% \end{center}
%    \caption{\textbf{Overview.} Our framework, Uni-renderer, empowers the generative model to function both as a renderer and an inverse renderer by approximating the rendering equation using a data-driven approach. Given intrinsic attributes, Uni-renderer generates photo-realistic images, functioning as a renderer. Furthermore, when provided with a single RGB image, it effectively decomposes the intrinsic properties, functioning as an inverse renderer.
%    \textbf{Top:} Uni-renderer generates smooth variations in response to different metallic and roughness values. Setting the roughness value to $1.0$ results in the ``dog" case, shown at the top, lacking specular highlights. Conversely, setting the metallic value to 1 makes the ``hat" case appear metallic.
%     \textbf{Bottom Left:} When functioning as an inverse renderer, Uni-renderer decomposes the intrinsic properties of a single RGB image. \textbf{Bottom Right:} Relighting demonstration of Uni-renderer given RGB images with spatially varying environment lighting.}

% \label{fig:teaser}
% \end{figure*}

\begin{strip}
    \centering
    \vspace{-3em}
    \centering
    \includegraphics[width=\textwidth]{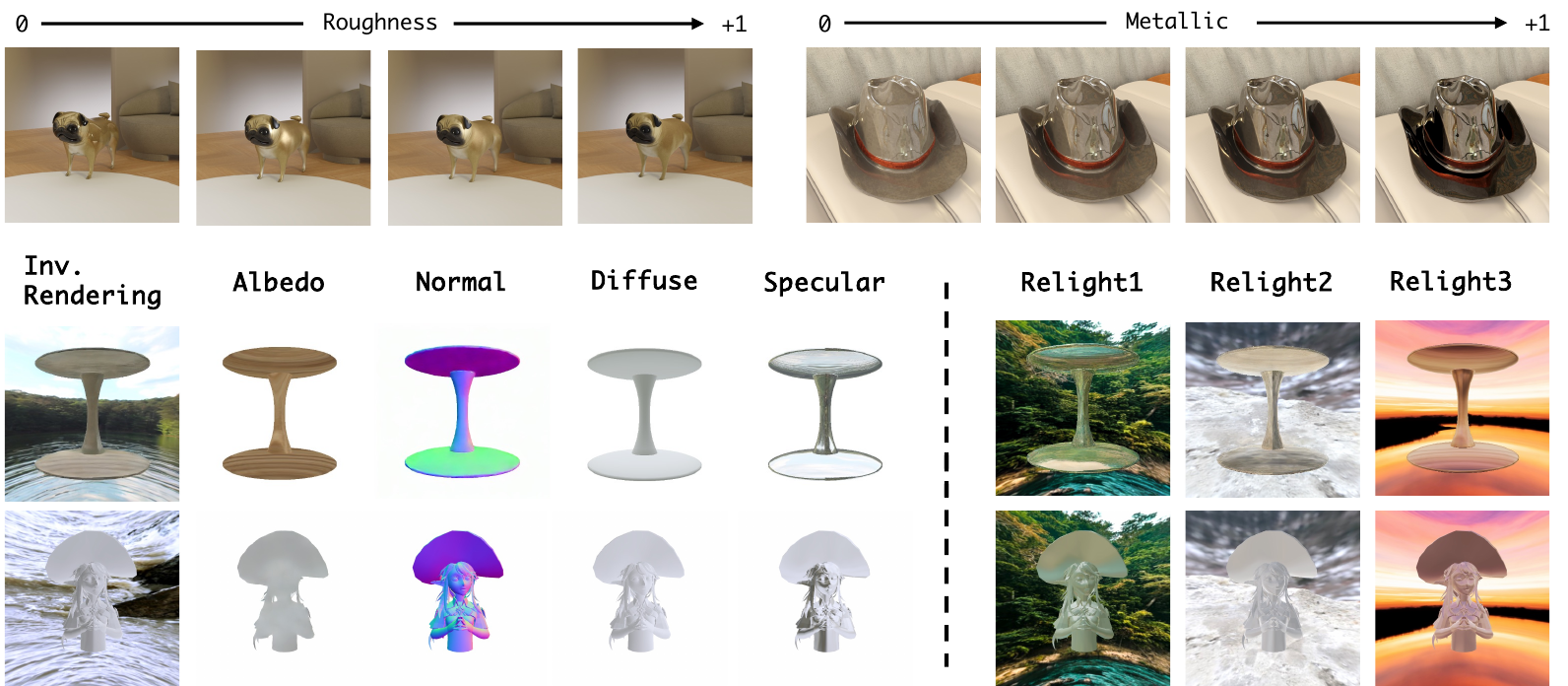}
    \vspace{-1.7em}
        \captionof{figure}{Our framework, Uni-renderer, empowers the generative model to function both as a renderer and an inverse renderer by approximating the rendering equation using a data-driven approach. Given intrinsic attributes, Uni-renderer generates photo-realistic images, functioning as a renderer. When provided with a single RGB image, it effectively decomposes the intrinsic properties, functioning as an inverse renderer.
   \textbf{Top:} Uni-renderer generates smooth variations as renderer. Setting the roughness value to $1.0$ results in the ``dog" case, shown at the top, lacking specular highlights. Conversely, setting the metallic value to 1 makes the ``hat" case appear metallic.
    \textbf{Bottom Left:} When functioning as an inverse renderer, Uni-renderer decomposes the intrinsic properties of a single RGB image. \textbf{Bottom Right:} Uni-renderer generates relighting results under different environment lighitngs.
    }
    \vspace{.5em}
    \label{fig:teaser}
\end{strip}
\renewcommand{\thefootnote}{}
\footnotetext{$\dag$ Equal contribution. $*$ Corresponding Author.}
\footnotetext{This project is supported by Quwan Technology.}
\begin{abstract}
Rendering and inverse rendering are pivotal tasks in both computer vision and graphics. The rendering equation is the core of the two tasks, as an ideal conditional distribution transfer function from intrinsic properties to RGB images. Despite achieving promising results of existing rendering methods, they merely approximate the ideal estimation for a specific scene and come with a high computational cost. Additionally, the inverse conditional distribution transfer is intractable due to the inherent ambiguity. To address these challenges, we propose a data-driven method that jointly models rendering and inverse rendering as two conditional generation tasks within a single diffusion framework. Inspired by UniDiffuser, we utilize two distinct time schedules to model both tasks, and with a tailored dual streaming module, we achieve cross-conditioning of two pre-trained diffusion models. This unified approach, named Uni-Renderer, allows the two processes to facilitate each other through a cycle-consistent constrain, mitigating ambiguity by enforcing consistency between intrinsic properties and rendered images. Combined with a meticulously prepared dataset, our method effectively decomposition of intrinsic properties and demonstrating a strong capability to recognize changes during rendering. We will open-source our training and inference code to the public, fostering further research and development in this area.

% for generating photorealistic images from 3D models and reconstructing scene attributes from images, respectively. Traditional methods, often based on Monte Carlo simulations, face challenges due to high computational costs and the inherently ill-posed nature of inverse rendering. We propose a unified parallel streaming diffusion framework that efficiently addresses both tasks within a single model. This framework adapts an off-the-shelf diffusion model to perform both rendering and inverse rendering in a cross-conditioned manner. Our model demonstrates robust disentanglement of intrinsic properties and. This novel approach not only streamlines traditional pipelines but also reduces dependency on auxiliary data, offering substantial advancements for real-world applications in game production, animation, and architectural visualization.
\end{abstract}

\section{Introduction}
    
Physically based rendering is crucial in computer vision and graphics for producing photorealistic 2D images from 3D models, textures, and lighting setups. This process is essential in animation \cite{winder2013producing}, and architectural visualization \cite{kvrivanek2018realistic}.
At its core, the rendering equation \cite{kajiya1986rendering} models the flow of light energy within a scene, offering an ideal conditional distribution transfer function for rendering. Monte Carlo light-transport simulations \cite{pharr2023physically}, utilizing path tracing \cite{lafortune1993bi}, are commonly employed to evaluate the rendering equation. However, its recursive tracing incurs significant computational demands, posing challenges for real-time implementation. Inverse rendering, which aims to deduce geometric, material, and lighting information from images, has long been studied due to its under-constrained nature \cite{grosse2009ground}. This approach allows the reconstructed 3D model to be directly integrated into rendering engines, thereby playing a critical role in downstream applications such as game production~\cite{lewis2002game}.

Recent advancements employ differentiable rendering for 3D representations \cite{chen2021dib, yao2022neilf} or focus on direct decomposition of intrinsics using 2D object priors \cite{guo2020materialgan, deschaintre2018single}. These approaches use differentiable renderer to integrate intrinsic properties into RGB images, facilitating loss computation against ground-truth images for optimization. 
For instance, MaterialGAN \cite{guo2020materialgan} leverages StyleGAN2 \cite{karras2020analyzing} to synthesize realistic material properties using a large-scale, spatially-varying material dataset \cite{deschaintre2018single}, while other works \cite{yi2023weaklysupervised, kocsis2024intrinsicimagediffusion, careagaIntrinsic, chen2024intrinsicanything} utilize diffusion models to achieve better results. Nevertheless, the ambiguity between geometry, materials, and lighting still hinders the effectiveness of decomposition. 
% However, these methods have not performed well due to the inherent ambiguity in transferring distributions from images to intrinsic properties; 
The mapping from observed images back to intrinsic properties is not one-to-one, leading to ambiguity and suboptimal performance. The most similar work to ours is RGB2X \cite{Zeng_2024}, which performs both forward and inverse rendering using two separate diffusion models. However, it fails to establish a connection between the two processes, treating them independently. Ambiguity issue still exists.

To alleviate this issue, we propose to jointly learn the two distributions together. By integrating rendering and inverse rendering into a single framework from a multi-task learning perspective, the two processes can facilitate each other. This joint learning allows us to use the inverted intrinsic properties to perform another cycle of rendering, effectively creating a cycle-consistent constrains.
 %framework. 
 This cycle rendering can mitigate the ambiguity problem by enforcing consistency between the intrinsic properties and the rendered images, leading to improved performance.
% such methods are merely approximations of the conditional distribution transfer module, requiring recursive ray tracing to re-evaluate the rendering equation on new scenes.

In this work,  we explicitly model the rendering process and inverse rendering as two conditional generation tasks. Inspired by UniDiffuser \cite{bao2023transformer}, we utilize two distinct time schedules to model both conditional generation tasks within a single diffusion pipeline. By cross-conditioning two pre-trained diffusion models through a dual streaming module, we achieve both rendering and inverse rendering in a unified framework.

%This design offers several advantages. First, physically based rendering is costly due to its recursive nature on per-object optimization, whereas diffusion models offer generalization across objects thereby boasting faster rendering speeds. Second, inverse rendering is an under-constrained problem because different combinations of geometry, materials, and lighting can produce identical rendered images, leading to a vast solution space. To address this challenge, we have prepared a dataset that systematically varies one of these three variables at a time while keeping the other two constant.  For example, by fixing the geometry and lighting, we can focus on changes in the rendered images caused by variations in materials. For instance, if a shiny material is expected to reflect light in a specific way but does not in the rendered image, it is eliminated as a possibility. 
 
To summarize, we introduce Uni-renderer which utilizes a pre-trained diffusion model to handle both rendering and inverse rendering in a unified framework. The key contributions of our method are as follows:

\begin{itemize}
    \item We propose a data-driven method to approximate the rendering equation. By modeling rendering and inverse rendering as two conditional generation tasks, we design a unified diffusion model with a tailored dual stream module for cross-condition generation.
    %introduce a unified diffusion model for simultaneous rendering and inverse rendering. Providing limited low-level material properties for training, the model efficiently learns to decouple the underlying relationships between these attributes and the resulting renderings.
    \item We render a synthetic dataset of different fine-grained material edits using 200K 3D objects with randomized intrinsic attributes by varying one of the rendering elements at a time while keeping the others constant.
    %We render a synthetic dataset of different fine-grained material edits using 300 3D objects and randomized intrinsic attributes.
    \item Extensive experiments validate the effectiveness of our method, achieving robust disentangling of intrinsic properties and demonstrating a strong capability to recognize changes in rendering.
\end{itemize}

\section{Related Work}
\begin{figure*}[t]
\centering
\makebox[\textwidth][c]{%
    \includegraphics[width=1.\linewidth]{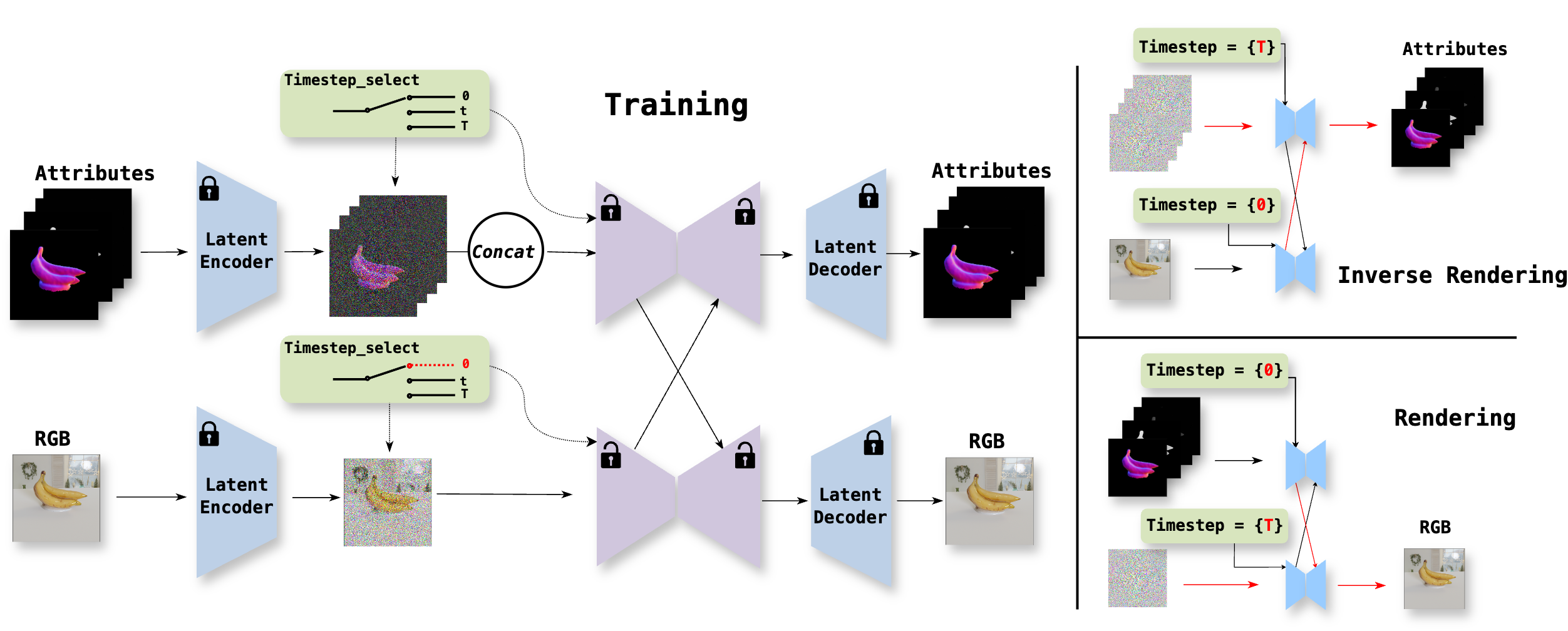}
}
   \caption{\textbf{The overview of our pipeline.} During training, both attribute and RGB images are input to a unified model with pre-trained VAE encoders. The timestep selector plays a crucial role by adjusting the timesteps for each branch. Specifically, it ensures that one branch (either the attribute or RGB) has a timestep of 0, while the other branch selects a timestep from $t \in [0, T]$. This mechanism allows our model to effectively learn the conditional distributions $q(\mathbf{x_0}|\mathbf{y_0})$, $q(\mathbf{y_0}|\mathbf{x_0})$ in alternating iterations. During rendering and inverse rendering, the corresponding conditions are input to the model with a timestep of 0, and the attributes/RGB images are generated through a sampled noise. (The VAE encoder and decoder are omitted for simplicity.) } 
\label{fig:pipeline}
\end{figure*}
\subsection{Rendering}
Rendering, the process of generating 2D images from 3D models, synthesizes raw data from a 3D scene—including geometry, materials, and lighting—using the rendering equation \cite{kajiya1986rendering}. This task is typically performed by rendering engines, such as Blender \cite{hess2013blender} and Unity \cite{haas2014history}, which serve as the technological intermediaries converting 3D models into 2D images or videos. However, achieving photo-realistic 2D images often involves recursive ray tracing \cite{shirley2008realistic} or path tracing \cite{lafortune1993bi}, techniques that closely approximate the rendering equation but are time-consuming. Although several GPU-based acceleration methods have been proposed \cite{kilgard2012gpu, everitt2003practical, heidrich1999realistic}, and existing engines have incorporated these tools, real-time rendering continues to pose significant challenges. In contrast to traditional rendering approaches, our method considers the diffusion model as an alternative renderer, which utilizes geometry, materials, and lighting as conditional cues to produce photo-realistic 2D images without the need for recursive tracing.

\subsection{Inverse Rendering}
Inverse rendering \cite{barron2014shape, nimier2019mitsuba}, the task of decomposing an image's appearance into intrinsic properties such as geometry, material, and lighting, remains a longstanding and severely ill-posed problem in computer vision and graphics. Some approaches employ differentiable renderers on 3D representations \cite{chen2021dib, kato2018neural} to directly optimize these properties using image losses. Advancements in volume rendering, as utilized in Neural Radiance Fields (NeRFs) \cite{mildenhall2021nerf} for 3D reconstruction from multi-view 2D images, have led most methods \cite{liang2023envidr, liu2023nero, zhang2022modeling, yao2022neilf, fan2023factored} to leverage multiple images for reconstruction and subsequent estimation of materials and lighting. Additionally, they require per-object optimization and struggle to generalize across different objects, limiting their efficiency in fast estimation. MaterialGAN \cite{guo2020materialgan} utilizes a StyleGAN2-based model for generating intrinsic properties and a renderer for novel view synthesis under specified lighting conditions. Similarly, SIC \cite{deschaintre2018single} employs an encoder-decoder network for single-image-based inverse rendering, where rendering loss is used for optimization. Nonetheless, these efforts predominantly focus on planar surfaces and still necessitate a renderer to integrate predicted attributes into the final rendered image. \cite{yi2023weaklysupervised, kocsis2024intrinsicimagediffusion, careagaIntrinsic, chen2024intrinsicanything, Zeng_2024} utilize diffusion models to achieve better results. However, these methods have not achieved optimal performance due to the inherent ambiguity involved in transferring distributions from images to intrinsic properties.
% add comparison
Our method effectively mitigates the ambiguity among intrinsic properties by introducing a dual-stream diffusion pipeline and combined with a meticulously prepared dataset.

\subsection{Diffusion Models}
% need to further talk about controlnet and such conditional generative models and \ unidiffusers
Diffusion Models \cite{ho2020denoising, song2020denoising, song2020score, nichol2021improved, rombach2022high} belong to the class of probabilistic generative models that progressively destruct data by injecting noise, and then subsequently learn to reverse this process in order to generate new samples.  Alchemist \cite{sharma2023alchemist} leverages a pre-trained diffusion model to achieve material control over a given object. \cite{vainer2024collaborative} adopts a collaborative diffusion pipeline to decompose physically-based rendering (PBR) material properties from RGB. However, these works focus either on rendering or inverse rendering and fail to yield a unified model to fit in real-world usages. Alternatively, we design a dual stream diffusion model as both a renderer and an inverse renderer. During the inverse rendering stage, it takes a single RGB image as input and disentangles all the intrinsic properties. For the rendering stage, all of the intrinsic properties including material, geometry, and lighting conditions will be taken and rendered via the same model.

\begin{figure*}[h]
\begin{center}
\includegraphics[width=1\linewidth]{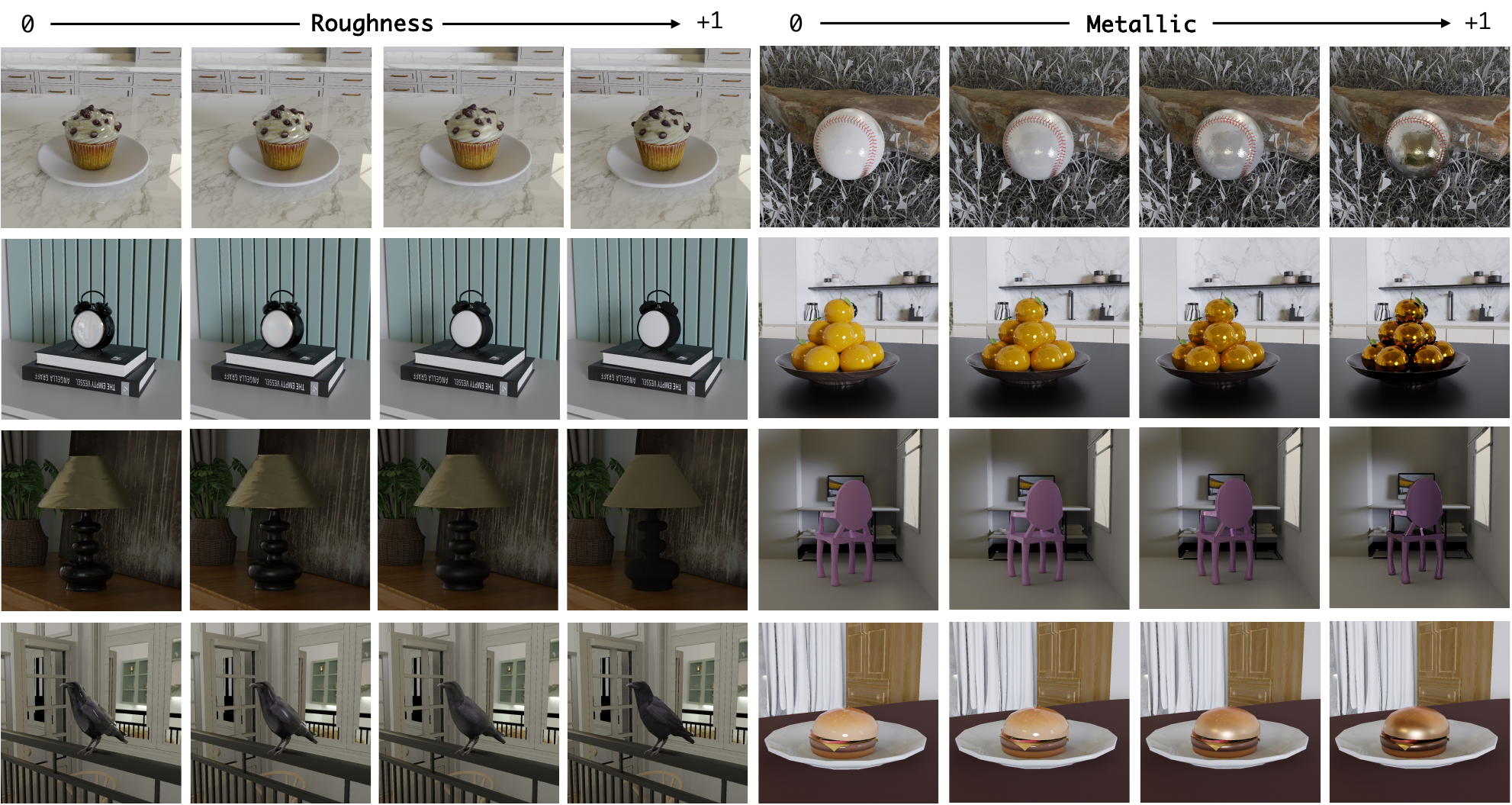}
\end{center}
\vspace{-0.5cm}
   \caption{We demonstrate smooth changes via rendering for different metallic and roughness strengths. The rendering is performed giving different combinations of the attributes. When the roughness value was set to 1, the cake and clock case shown in the top left are without specular highlights. When the metallic value was set to 1, the orange and baseball cases appeared to be metallic and revealing object illumination. \textbf{Best viewed in color.}} 
\label{fig:qualitative_1}
\end{figure*}

\section{Approach}
Given a single RGB image $\mathbf{I}$, inverse rendering aims at decomposing it into intrinsic attributes including metallic $m$, roughness $r$, albedo $\mathbf{a}$, surface normal $\mathbf{n}$, specular lighting $\mathbf{s}$ and diffuse lighting $\mathbf{d}$, which is formulated as ${\{m,r,\mathbf{a}, \mathbf{n}, \mathbf{s}, \mathbf{d}\} = \mathrm{Inverse Renderer(\mathbf{I})}}$. For later reference, we denote the combination of attributes as $\mathbf{C}$.
Renderer takes attributes as input and renders them into RGB image denoted as ${ \mathbf{I}= \mathrm{Renderer(\mathbf{C})}}$. We formulate a dual stream diffusion model as renderer and inverse renderer simultaneously. 
We start by briefly revisiting physically based rendering (PBR) and Unidiffuser \cite{bao2023transformer} in Section \ref{Preliminaries}. Next, we introduce our dual stream diffusion framework in Section \ref{Network}. We elaborate on the data preparation in Section \ref{data Preparation}.

\subsection{Preliminaries} \label{Preliminaries}

\vspace{0.1cm}
\textbf{Physically Based Rendering. } Physically-based rendering is a computer graphics approach that seeks to render images by modeling the interaction between lights and materials in the real world. At its core, the rendering equation \cite{kajiya1986rendering} describes the light energy flow in the scene, formulated as 

\begin{equation}\label{rendering equation}
L(\textbf{x},\bm{\omega_o}) = \int_{\Omega}f(\textbf{x}, \bm{\omega_o}, \bm{\omega_i}) L_i(\textbf{x}, \bm{\omega_i})  (\bm{\omega_i}, \textbf{n}) d\bm{\omega_i},
\end{equation}
where $\bm{\omega_o}$ is the viewing direction of the outgoing light, $f$ is the BRDF properties, $L_i$ is the incident light of direction $\bm{\omega_i}$ sampled from the upper hemisphere $\Omega$, and $\textbf{n}$ is the surface normal. 
The rendering equation is the key to handle both rendering and inverse rendering problems. Given intrinsic properties, the rendering equation generates photo-realistic images. Given an image, a rendering equation can also be used to optimize intrinsic properties.

\vspace{0.1cm}
\noindent
\textbf{Unidiffuser.}
% UniDiffuser \cite{bao2023transformer}, a unified diffusion model that models all distribution at the same time, it injects noise $\epsilon^x$ and $\epsilon^y$ to a set of paired image-text data \textbf$(x_0, y_0)$ and generates noisy data $\textbf{x_t^x}$ and \textbf{$y_{t^y}$}, where $0 \leq t_x, t_y \leq T$ represent two individual timesteps.  It then trains a joint noise prediction network $\epsilon_{\theta}(\textbf{x_{t^x}}, \textbf{y_{t^y}}, t_x, t_y)$ to predict the noise $\epsilon^x$ and $\epsilon^y$ by minimizing the mean squared error loss:
UniDiffuser \cite{bao2023transformer}, a unified diffusion model that models all distributions at the same time, injects noise $\epsilon^x$ and $\epsilon^y$ to a set of paired image-text data $(\mathbf{x_0}, \mathbf{y_0})$ and generates noisy data $\mathbf{x_t^x}$ and $\mathbf{y_t^y}$, where $0 \leq t_x, t_y \leq T$ represent two individual timesteps. It then trains a joint noise prediction network $\epsilon_{\theta}(\mathbf{x_{t^x}}, \mathbf{y_{t^y}}, t_x, t_y)$ to predict the noise $\epsilon^x$ and $\epsilon^y$ by minimizing the mean squared error loss:

\begin{equation}\label{unidiffusers_mse}
\mathop{\mathbb{E}}_{\mathbf{\epsilon^x}, \mathbf{\epsilon^y}, \mathbf{x_0}, \mathbf{y_0}} \left[ \left\| \left[ \mathbf{\epsilon^x}, \mathbf{\epsilon^y} \right] - \epsilon_{\theta}(\mathbf{x_{t^x}}, \mathbf{y_{t^y}}, t_x, t_y) \right\|^2 \right],
\end{equation}

% \begin{equation}\label{unidiffusers mse}
% \mathop{\mathbb{E}}_{\epsilon^x, \epsilon^y, x_0, y_0}[||[\epsilon^x, \epsilon^y]-\epsilon_{\theta}(x_{t^x}, y_{t^y}, t_x, t_y)||^2],
% \end{equation}
By predicting $\epsilon_{\theta}(\mathbf{x_{t^x}}, \mathbf{y_{t^y}}, t_x, t_y)$ for any $t_x$ and $t_y$, UniDiffuser learns all distributions related to $(\mathbf{x_0}, \mathbf{y_0})$. This includes all conditional distributions: $q(\mathbf{x_0}|\mathbf{y_0})$, $q(\mathbf{y_0}|\mathbf{x_0})$, and those conditioned on noisy input, for example $q(\mathbf{x_0}|\mathbf{y_{t^y}})$ and $q(\mathbf{y_0}|x_{t^x})$, for $0 \leq t_x, t_y \leq T$. Learning a conditional distribution can be seen as learning a distinct task. From a multitask learning perspective, due to limited network bandwidth, learning many tasks simultaneously (i.e., fitting all distributions to a single network) may result in task competition or task conflict, ultimately leading to sub-optimal performance. For rendering and inverse rendering, we exclusively modeled the two conditional distributions $q(\mathbf{x_0}|\mathbf{y_0})$, $q(\mathbf{y_0}|\mathbf{x_0})$ to resolve the aforementioned issue and enhance the performance. 
\begin{figure*}[h]
\begin{center}
\includegraphics[width=1.\linewidth]{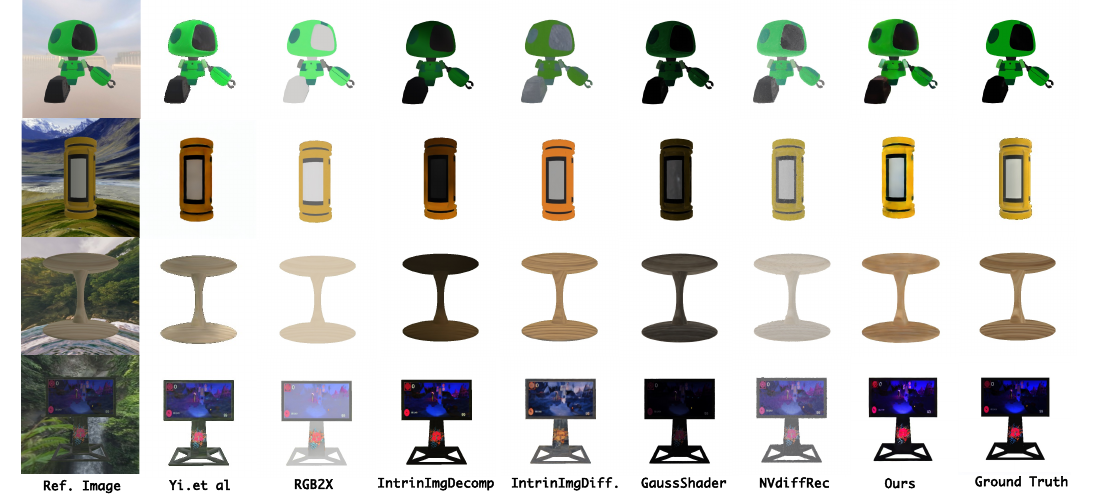}
\end{center}
   \caption{\textbf{Albedo comparison.} Albedo Comparison of Uni-Renderer with baseline methods. We compared 4 learning-based methods and 2 optimization-based methods. Among all, Uni-renderer yields the most realistic results.  \textbf{Best viewed in color.}} 
\label{fig:albedo}
\end{figure*}

\subsection{Uni-Renderer} \label{Network}
% pipeline chart goes this and talks about how the model works briefly 
Our design incorporates a pre-trained RGB image model and a PBR model, the two were tightly linked to one another via a dual streaming technique aligning the RGB model’s information with the PBR model’s information. The overall framework is shown in Figure~\ref{fig:pipeline}. Next, we will discuss each sub-module in the pipeline in detail. 

\vspace{0.1cm}
\noindent
\textbf{Latent Preparation.} Recent advancements have benefited greatly from the pre-trained VAE \cite{rombach2022highresolution}, which downsample an RGB image into lower dimensional latent. We used separate VAE for encoding $[{m}$, ${r}]$, albedo $\mathbf{a}$, surface normal $\mathbf{n}$ and environment lighting $\mathbf{s}$, $\mathbf{d}$ before we channel-wisely concatenate them and feed them into the model. It is worth noting that attributes like metallic and roughness, often come with a scalar value, and cannot be directly encoded. Thus, we generate additional masks to binarize the two values into gray-scale images.  Instead of having two separate VAE for $\mathrm{m}$ and $\mathrm{r}$, We formed channel triplets $a_{\text{material}}$, consisting of [$m$, $r$, $\bold{m}$] and process those with the RGB VAE, where $\bold{m}$ is the mask This design allows us to reduce the bandwidth overhead within the diffusion model and further improve generation quality.   

\vspace{0.1cm}
\noindent
\textbf{Conditional Distributions Modeling.}
% how to model the two conditional distribution need to be discuss in preliminaries 
To achieve rendering and inverse rendering within a single model, we borrowed ideas from \cite{bao2023transformer}, and adopted two different timesteps $t_{\text{attributes}}$, $t_{\text{RGB}}$ from three timesteps choices $\{0, t, T\}$ to train a joint noise prediction network. It is worth noting that since our unified model only requires modeling two conditional distributions, we ruled out the choices where both $t_{\text{attributes}}$ and $t_{\text{RGB}}$ take on $t$ or $T$ and make sure there is always a timestep equal to $0$. Formally,
\begin{equation} 
\left( t_{\text{attributes}},\ t_{\text{RGB}} \right) = \begin{cases} \left( 0,\ \tilde{t} \right) & \text{during rendering} \\ \left( \tilde{t},\ 0 \right) & \text{inverse} \end{cases},
\end{equation}

where

\begin{equation} 
\tilde{t} = \begin{cases} t & \text{with probability } p \\ T & \text{otherwise}
\end{cases}.
\end{equation}
The pseudo-code for generating timesteps can be found in supplementary material.
% provide additional algorithm 

\vspace{0.1cm}
\noindent
\textbf{Dual Stream Diffusion.} \label{Dual}
We designed a dual diffusion network to facilitate the unified model and adopt the $\mathbf{x}_0$-prediction in the diffusion process.
\textbf{Training}: During training,  as shown in Figure~\ref{fig:pipeline}, the upper branch of the model dealt with the intrinsic attributes only, leaving only RGB for the lower branch. The timestep selection module will determine if a rendering/inverse rendering iteration has occurred at a probability $p$. During a rendering iteration, the clean attributes $\mathbf{C}$ will be used as conditions to denoise the noisy RGB image $\mathbf{I}$, the same will be applied to the inverse rendering iteration except that the noisy attributes are being fed to the model.
\textbf{Inference}: At the inference stage, we provided attributes as conditions to the model and gradually denoise the RGB, and vice versa. The effectiveness of the dual stream framework is demonstrated in Sec~\ref{ablation}. 

\vspace{0.1cm}
\noindent
\textbf{Cycle-Consistent Constrain.} \label{CYCLE}
To alleviate the inherent ambiguity problem, we introduce a cycle-consistent constraint within our unified framework, where its incorporation is straightforward with our pipeline. During training, we use the model's predicted inverse results to perform an additional cycle of rendering. The re-rendered outputs are then utilized in the loss calculation to further optimize the inverse rendering process. Formally, 
\begin{equation}
\mathcal{L} = \mathbb{E}_{\mathbf{x}_0, \boldsymbol{\epsilon}, t} \left[ \left\| \mathbf{x}_0 - \hat{\mathbf{x}}_0\left( \hat{\mathbf{x}}_\text{rgb}, t, \mathbf{C} \right) \right\|^2 \right],
\end{equation}
where $\hat{\mathbf{x}}_0\left( \mathbf{x}_t, t, \mathbf{C} \right)$ is the model's prediction of the original rendering output $\mathbf{x}_0$, conditioned on the noisy input $\hat{\mathbf{x}}_\text{rgb}$, timestep $t$, and conditioning attribute information $\mathbf{C}$.
The effectiveness of the cycle-consistent constrain is demonstrated in Sec~\ref{ablation}.

\subsection{Data Preparation} \label{data Preparation}

We prepare our training data with 3D data from Objaverse \cite{deitke2023objaverse}, consisting of synthetic 3D assets. We sampled 200K assets from Objaverse. For each 3D asset in Objaverse, we rendered the 2D images, and materials map (i.e., metallic, roughness, albedo) by changing the metalness and roughness (range from 0 to 1 with step 0.1). We also randomly select a 2D environment map from a pool of 20K natural scenes from a subset of LHQ-1024 \cite{lhq1024_dataset} and format them into an RGB style for providing lighting. For each object, we have 121 pairs with different metallic, roughness and lighting. The rendered images are of resolution 1024 $\times$ 1024,  and the camera poses are fixed in the front of the object.
We also rendered 100 objects that were unseen during training to test the model generalization ability.

\begin{figure}[t]
\begin{center}
\includegraphics[width=1\linewidth]{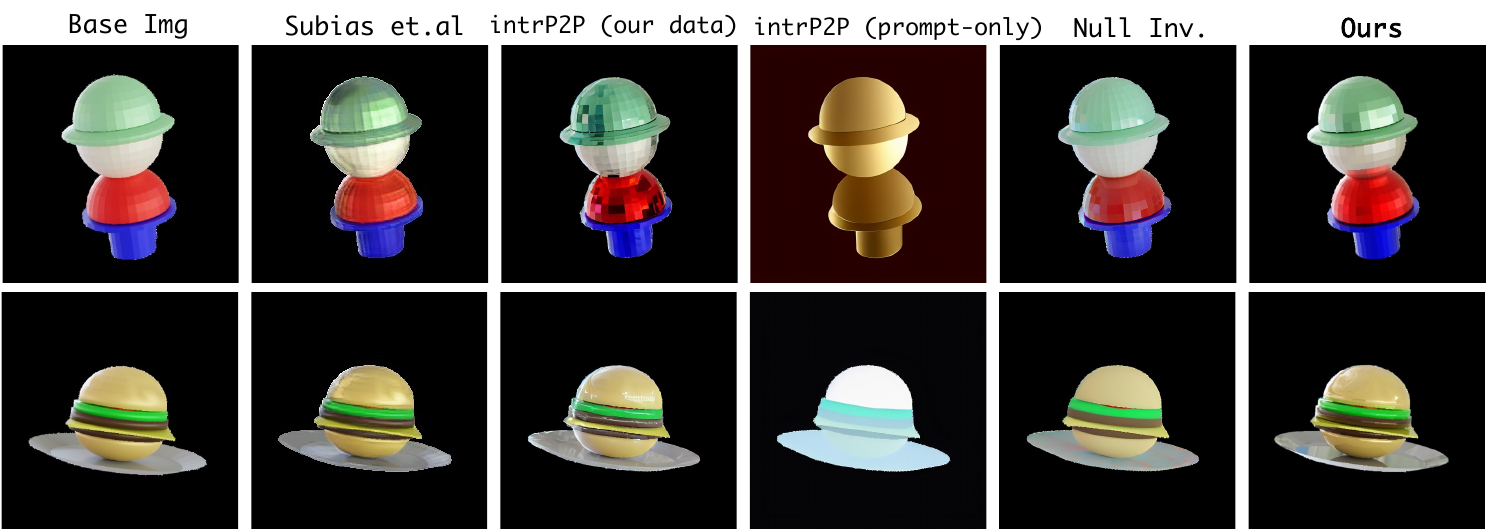}
\end{center}
   \caption{\textbf{Qualitative comparison.} Rendering Comparison of Uni-Renderer with baseline methods. The base images are with a metallic value of 0.5. The comparison is made with a higher metallic value of 1.0.  \textbf{Best viewed in color.}} 
\label{fig:qualitative_comp}

\end{figure}
\section{Experiments}
%qualitative for metallic and roughness change, env inverse
We present qualitative and quantitive analysis in Section~\ref{quali_1} to demonstrate the rendering capabilities of our model. The inverse rendering results can be found in Section~\ref{comparison}.

\noindent

\begin{table}[h]
\small
   \centering
   \vspace{-0.3cm}
   \caption{\textbf{Quantitative results on rendering.} Metrics for the prompt-only InstructPix2Pix* trained on our data, Subias et.al and our proposed method computing the PSNR, and LPIPS. }
    \label{rendering}  
    \resizebox{.45\textwidth}{!}{
    \begin{tabular}{c|cc|cc}
         \hline
            \multirow{2}*{Methods}& \multicolumn{2}{c|}{Metallic} & \multicolumn{2}{c}{Roughness}  \\ \cline{2-5} 
            & PSNR$\uparrow$   &LPIPS$\downarrow$  &PSNR $\uparrow$   &LPIPS$\downarrow$   \\  \hline\hline
          $\text{InPix2Pix*}$ \cite{brooks2023instructpix2pix} &24.25 &0.1032 &24.43 &0.1056    \\  
          $\text{Subias et.al}$ \cite{subias2023inthewild}   &28.09 & 0.0954 &28.13 &0.0817   \\
          \hline
          $\text{Ours}$  &\textbf{30.72} &\textbf{0.0763} &\textbf{31.68} &\textbf{0.0695}  \\
          $\text{Ours w/o unified}$  &27.33 &0.0932 &29.12 &0.0987  \\
          $\text{Ours w/o re-render}$  & 28.72 & 0.0824 & 28.93 & 0.0833  \\
        \hline
    
    \end{tabular}}
    \vspace{-0.1in}
\end{table}

\subsection{Rendering} \label{quali_1}
In Figure~\ref{fig:qualitative_1}, we demonstrate the effectiveness of our model in generalizing rendering techniques by utilizing different combinations of material attributes. Using our dual streaming pipeline, we start off by passing a reference image to perform inverse rendering. Then update the metallic, roughness value and finally re-render into an RGB image with updated materials. 

\vspace{0.1cm}
\noindent
\textbf{Roughness.} % explains how roughness related to PBR rendering
As the roughness increases, the output demonstrates the elimination of specular highlights, replaced by an estimate of the base albedo. Conversely, reducing the roughness amplifies the highlights, as evidenced in the cases of the crow and the lamp in Figure~\ref{fig:qualitative_1}. 

\vspace{0.1cm}
\noindent
\textbf{Metallics.} % explains how metallics change, the albedo plays a less important role
The increase of the metallic strength in both the baseball and oranges in Figure~\ref{fig:qualitative_1} reduces the significance of the base albedo and enhances the lighting effects on the surfaces. In contrast, reducing the metallic strength reverses this effect.

\vspace{0.1cm}
\noindent
\textbf{Analysis.}
We compare our model to other material editing baselines: Subias et al. \cite{subias2023inthewild}, Prompt-to- Prompt with Null-text Inversion (NTI) \cite{mokady2022nulltext}, and InstructPix2Pix \cite{brooks2023instructpix2pix} in Figure~\ref{fig:qualitative_comp}.  We fine-tuned the InstructPix2Pix prompt-based approach with our dataset. Subias et al.’s method results in exaggerated material changes as their objective is perceptual, not physically-based material rendering. Null-text inversion and InstructPix2Pix trained on our dataset with a prompt-only approach, they either incorrectly modified the background or produces minimum changes. Our method renders the photo-realistic results, introducing the specular highlights while retaining the geometry and illumination effects. Additionally, the quantitative results with the aforementioned baselines are shown in Table~\ref{rendering}.

\begin{figure*}[t]
\begin{center}
\includegraphics[width=0.95\linewidth]{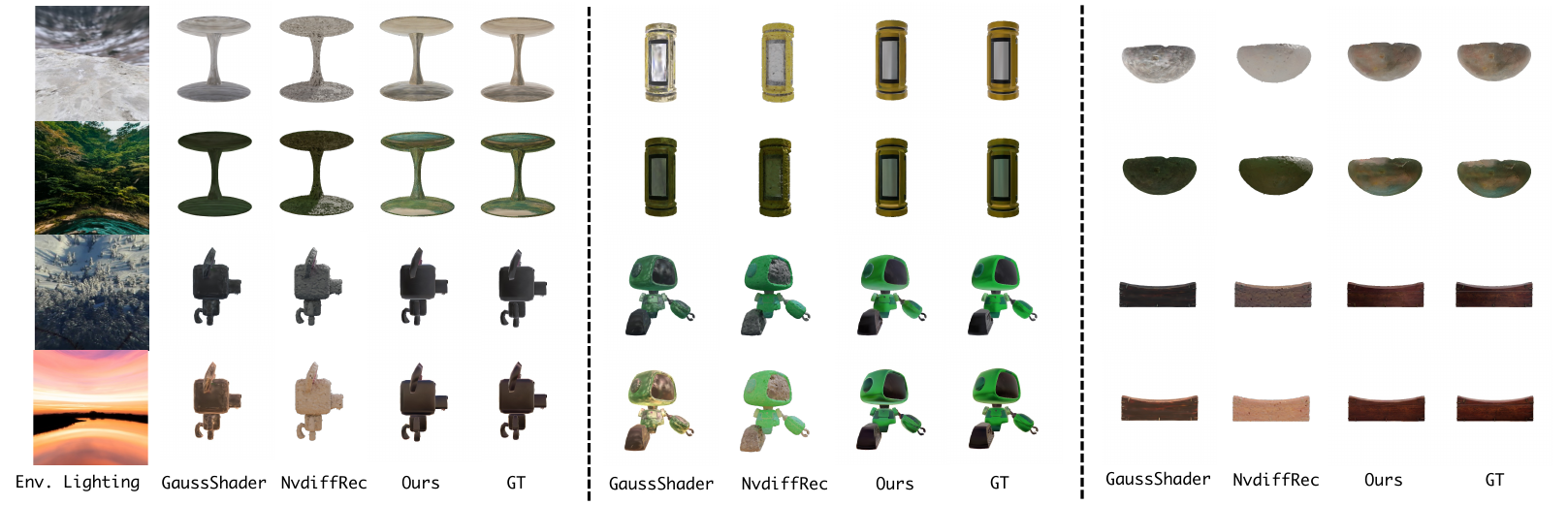}
\end{center}
\vspace{-0.4cm}
   \caption{\textbf{Relighting Comparsion} The relighting comparison is performed on validation objects. We first inverse render the input RGB to acquire the intrinsics, and then we updated the lighting information to get the relighting results. The leftmost column is the reference environment lighting.  \textbf{Best viewed in color.}} 
\label{fig:relighting}
\end{figure*}

\subsection{Inverse Rendering} \label{comparison}
We use Peak Signal-to-Noise Ratio (PSNR), Structural Similarity Index Measure (SSIM), and Learned Perceptual Image Patch Similarity (LPIPS) as metrics to evaluate the image quality of the albedo, relighting images, and novel view synthesis. We use Cosine Similarity for normal evaluation. Roughness and metallic estimation are evaluated using Mean Squared Error (MSE). Detailed configurations of the comparisons are discussed in the supplementary file.

\noindent
\subsubsection{Lighting Estimation}

% qualitative comparison for normal and albedo, rendering comparison with editing methods
We compare our model with NVdiffRec~\cite{Munkberg_2022_CVPR}, GaussianShader~\cite{jiang2023gaussianshader} in terms of specular and diffuse lighting decomposition, the results can be found in Figure~\ref{fig:relighting} and in Table~\ref{tab:table2}. In Figure~\ref{fig:relighting}, we demonstrate the relighting ability of our proposed model given different environmental lighting. To achieve this, we updated the specular, diffuse, and environment maps of the given objects. The rendered outputs under diverse lighting scenarios show consistent adjustments, preserving the material attributes while accurately reflecting the changes in illumination. We have demonstrated a superior relighting ability compared to other baselines. 

\begin{table}[t!]
\centering
\caption{
\textbf{Quantitative results on inverse rendering.}
For quantitative comparison, we have included both data-driven methods and optimization-based baselines. ``Ours w/o unified" is trained without the unified framework, and ``Ours w/o constrain" is trained without the cycle-consistent constrain term. 
}
\vspace{-0.3cm}
\resizebox{.5\textwidth}{!}{
\setlength{\tabcolsep}{1pt} %
\renewcommand{\arraystretch}{1.2} %
\begin{tabular}{ccccccccccc} 
\toprule
                                                                        &  & \multicolumn{3}{c}{Albedo} &  & Metallic &  & Roughness  &  & Normal  \\ 
\cline{3-5}\cline{7-7}\cline{9-9}\cline{11-11}
Method                                                               &  & PSNR$\uparrow$ & SSIM$\uparrow$ & LPIPS$\downarrow$   &  & MSE$\downarrow$        &  & MSE$\downarrow$                          &  & cos-simi$\uparrow$        \\ 
\midrule
Yi \etal  \cite{yi2023weaklysupervised}                                                   
 & & 20.93 & 0.8960 & 0.1024 &  & -  & & -  & & -

    \\
IntrinsicAny.  \cite{chen2024intrinsicanything}                                                   
 & & 22.67 &  \textbf{0.9218}& 0.0633 &  & -  & & -  & & -
\\
Wonder3d \cite{long2023wonder3d}
 & & - & - & - &  & -  & & -  & & 0.834 \\
Intrin.ImgDecomp. \cite{careagaIntrinsic}     
 & & 21.91 & 0.8934  & 0.0659 & & -  & & -  & & - \\
Intrin.ImgDiff. \cite{kocsis2024intrinsicimagediffusion}    
 & & 21.83 & 0.9060 & 0.0632 &  & 0.1920  & & 0.1315  & & - \\
 RGB2X. \cite{Zeng_2024}    
 & & 18.15 & 0.8829 & 0.0851 &  & -  & & -  & & 0.871 \\
 
\midrule
NvDiffrec                  \cite{Munkberg_2022_CVPR}                                             &  
  & 13.56 & 0.8695 & 0.1243 &  & 0.3164  & & 0.2956  & & 0.631 \\
GaussianShader \cite{jiang2023gaussianshader}   &  
  &  16.55   & 0.8640 & 0.0906 &  & 0.3421  & & 0.3714  & & 0.908      \\
\midrule
Ours                                                             
 & & \textbf{23.20} & 0.9182 & \textbf{0.0532} &  & \textbf{0.1182}	  & & \textbf{0.1037}  & & \textbf{0.928} \\
Ours w/o unified                                                                
 & & 18.62 & 0.8846 & 0.0833 &  &  0.1632	  & & 0.1391  & & 0.867 \\
Ours w/o constrain                                                             
 & & 21.20 & 0.8934 & 0.0602 &  &  0.1391  & & 0.1304  & & 0.922 \\

\bottomrule

\end{tabular}
}
\label{tab:table1}
    
\end{table}
\begin{table}[t!]
\centering
\caption{
\textbf{Quantitative results on relighting.}
We have compared the inverse rendering results on lighting estimation in our left-out test set, ``Ours w/o unified" is trained without the unified framework, and `Ours w/o constrain" is trained without the cycle-consistent constrain. 
}
\vspace{-0.3cm}
\resizebox{.5\textwidth}{!}{
\setlength{\tabcolsep}{2pt} %
\renewcommand{\arraystretch}{1.4} %
\begin{tabular}{ccccccccccccc} 
\toprule
                                                                        &  & \multicolumn{3}{c}{Specular} &  & \multicolumn{3}{c}{Diffuse}  &  & \multicolumn{3}{c}{Relighting}  \\ 
\cline{3-5}\cline{7-9}\cline{11-13}
Method                                                               &  & PSNR$\uparrow$ & SSIM$\uparrow$ & LPIPS$\downarrow$   &  & PSNR$\uparrow$ & SSIM$\uparrow$ & LPIPS$\downarrow$        &  & PSNR$\uparrow$ & SSIM$\uparrow$ & LPIPS$\downarrow$                                \\ 
\midrule

NvDiffrec                  \cite{Munkberg_2022_CVPR}                                              
 & & 14.12 & 0.8822 & 0.1782 &  & 15.20 & 0.8834 & 0.1833  &  & 21.99 &  0.8850 &  0.0834      \\
GaussianShader \cite{jiang2023gaussianshader}    
& & 16.82 & 0.8912 & 0.0982 &  & 17.99 & 0.8969 & 0.0831  &  & 26.47 & 0.8826 & 0.0822       \\
\midrule
Ours                                                                
 & & \textbf{22.71} & \textbf{0.9366} & \textbf{0.0498} &  & \textbf{23.15} & \textbf{0.9694} & \textbf{0.0373}  &  & \textbf{30.84}  & \textbf{0.9032} & \textbf{0.0763}       \\
Ours w/o unified                                                            
 & & 21.95 & 0.8816 & 0.0554 &  & 21.92 & 0.8934 & 0.0495  &  & 26.95 & 0.8832 & 0.0824       \\
Ours w/o constrain                                                            
 & & 22.07 & 0.9027 & 0.0513 &  & 22.82 & 0.9259 & 0.0408  &  & 28.12 & 0.8934 & 0.0924       \\

\bottomrule

\end{tabular}
}
\label{tab:table2}
    
\end{table}
\begin{figure*}[t]
\begin{center}
\includegraphics[width=1.\linewidth]{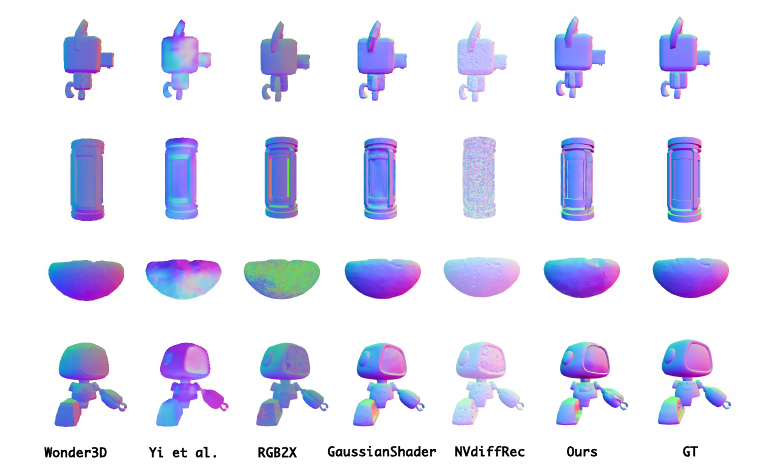}
\end{center}
\vspace{-0.3cm}
   \caption{\textbf{Normal Comparison.} Normal Comparison of Uni-Renderer with others methods.  \textbf{Best viewed in color.}} 
\label{fig:normal}
\end{figure*}

\noindent %确保接下来的内容不会缩进
\subsubsection{Materials and geometry}

\noindent
\textbf{Surface Normal.}
 We present the surface normal comparison results in Figure~\ref{fig:normal}. Our model yields the best performance in normal estimation when compared to Wonder3D \cite{long2023wonder3d}, RGB2X \cite{Zeng_2024}, and Yi et al, \cite{yi2023weaklysupervised}. 

\vspace{0.1cm}
\noindent
\textbf{Albedo.}
For albedo estimation, we have compared 2 optimization-based methods and 6 data-driven methods in terms of PSNR, SSIM and LPIPS. Figure~\ref{fig:albedo} presents the qualitative comparison for albedo estimation, and the quantitative comparison can be found in Table~\ref{tab:table1}. Our method significantly outperforms other methods.

\vspace{0.1cm}
\noindent
\textbf{Metallic and Roughness.}
Since existing data-driven methods do not include metallic and roughness estimation, we have compared the roughness and metallic estimation with two optimization-based inverse rendering method, including NVdiffRec~\cite{Munkberg_2022_CVPR} amd GaussianShader~\cite{jiang2023gaussianshader}, the quantitative results can be found in Table~\ref{tab:table1}. More visualization cases can be found in the supplementary material.

\subsubsection{Real-World Inversing}
\begin{figure}[h]
\begin{center}

\includegraphics[width=1.\linewidth]{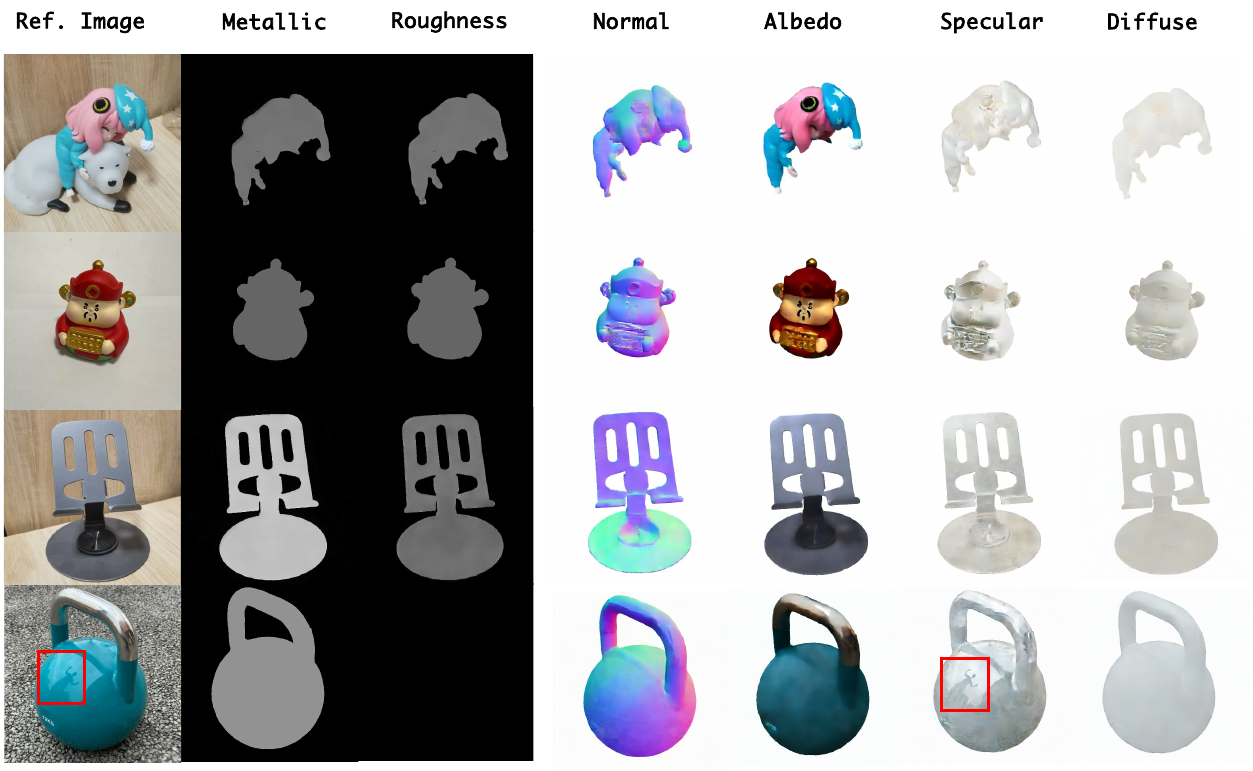}
\end{center}
\vspace{-0.4cm}
   \caption{\textbf{Real World Inversing.} We present different real-world inversing cases under different lighting conditions.  \textbf{Best viewed in color.}} 
\label{fig:real_world}
\end{figure}

Despite being trained on synthetic data, our model is capable of performing inverse rendering in real-world cases. As shown in Figure~\ref{fig:real_world}, the ``Phone stand'' case has a more metallic appearance while maintaining some roughness. It is under a white-to-yellow light source thereby showing a white-to-yellow color in lighting. 
Our model also excels at recovering high-frequency ambient lighting from the image as specular lighting, as demonstrated in the ``kettle" case.

\subsection{Ablation Study} \label{ablation}
\textbf{Dual Stream Diffusion.} To demonstrate the effectiveness of our dual stream diffusion framework, we compared results with two separate diffusion models for performing rendering and inverse rendering. Combining the two models in a single pipeline for both rendering and inverse can benefit each other, as shown in Table~\ref{tab:table1}, \ref{tab:table2} where the jointly trained model performance prevails. From a multi-task learning perspective, the unified pipeline benefits each other by sharing the knowledge through model weights.

\vspace{0.1cm}
\noindent
\textbf{Cycle-Consistent Constraint.} We also conducted experiments to show the effectiveness of the cycle-consistent constraint. The constrain can mitigate the ambiguity problem by enforcing consistency between the intrinsic properties and the  images, leading to improved performance. The results in Table~\ref{tab:table1}, \ref{tab:table2} demonstrated the inverse rendering quality improves with the constrain term being included.
 
% scaling law comparison 

% Attribute Guidance and image Guidance, how this is done with extra t = 999 and some visual comparision for different and (?quntitative comparison)
% \textbf{Model Scalability}
% We conducted a series of scalable experiments with varying numbers of input objects. As depicted in Figure~\ref{scaling_law}, the purple, green, and blue curves represent 30, 100, and 300 objects used for training, respectively. Rendering performance was evaluated using LPIPS\cite{nimier2019mitsuba}. At approximately 25K iterations, the rendering performance for the 30-object scenario reached its peak, indicated by the lowest LPIPS value. Subsequently, performance declined as LPIPS values increased, suggesting the model began to overfit, leading to unrealistic results. Similar trends were observed for the 100-object and 300-object scenarios, each demonstrating lower LPIPS values and longer durations before overfitting occurred. These results indicate our model's scalability, suggesting potential improvements in generalization with increased training data.

\section{Conclusion}

We present a method that allows photo-realistic rendering and inverse rendering in a unified framework. Our method demonstrates strong performance in real-world cases, achieving photo-realistic rendering and inverse rendering that generalizes well beyond synthetic training data. Despite this success, there are still limitations when applying our approach to certain real-world scenarios, primarily due to the domain gap between synthetic training data and real-world images. This gap can lead to challenges in accurately handling complex or unfamiliar objects and environments, where the synthetic-to-real adaptation is less effective. To address this, future work will focus on incorporating more real-world data into our training process. By expanding our dataset with real-world examples, we aim to enhance the model's ability to bridge the synthetic-real domain gap, enabling better generalization and higher-quality rendering results across diverse real-world scenes.

{
    \small
    \bibliographystyle{ieeenat_fullname}
    \bibliography{main}
}
\clearpage
\setcounter{page}{1}

\maketitlesupplementary
\section{Appendix / supplemental material}
In this supplementary, we will first discuss the detailed network architecture and the detailed algorithm for calculating different timesteps for reducing the tasks spaces. Then we will provide a description of the configurations used for baseline comparison.  
We also include more qualitative cases to demonstrate the capacity of our framework to perform smooth rendering and inverse rendering. 

\subsection{Dual Stream Diffusion}
The design of our framework involves two parallel stable diffusions. The upper branch takes in channel-concatenated attributes. Its UNet input $``Conv\_in"$ and output $``Conv\_out"$ layers are modified and extended to 24 channels for the corresponding input and output latents. The lower branch remains unchanged. The communication between the upper and lower branch are implemented through a cross-connected manner. We first take the intermediate feature $``mid\_block\_res\_samples"$ from the upper encoder and add it to the lower decoder through a zero convolution layer. We do the same for the lower encoder. Such design effectively enables the communication between two stable diffusions in a cross-conditioned manner. The introduction of the zero convolution layer maintains the pertaining weight not get disrupted during training.

\subsection{Model Training}
During training, we adopted the x0 prediction into our loss calculation. It effectively helps to solve the channel bandwidth overhead problem. The training of diffusion models is performed on eight A800 GPUs, with a batch size of 4, a learning rate of $1e-5$, and a total training iteration number of 150,000. We utilize the Adam optimizer with $adam\_beta1$ and $adam\_beta2$ equal to 0.8 and 0.999, respectively.

\subsection{Modeling conditional distributions with Two Timesteps}
To achieve rendering and inverse rendering within a single model, we introduced a reduced timestep strategy to eliminate redundant tasks and thereby speed up convergence time and generation quality. In algorithm 1, we showed the pseudo-code for generating different timesteps. 

\begin{algorithm}
\caption{Compute time steps matrix}
\begin{algorithmic}[1]
\Require{$len\_t$} \Comment{Length of timesteps matrix, defaults to $2$ in our case.}
\Require{$num\_timesteps$} \Comment{Number of timesteps, ranging from 0 to $T$.}
\Require{$bs$} \Comment{Batch size}
\State $timesteps \gets \text{initialize a zero matrix of size } len\_t \times bs$
\State $idx \gets \text{random integer from } 0 \text{ to } len\_t-1$
\State $all\_t[idx] \gets \text{random integers from } 0 \text{ to } num\_timesteps-1 \text{ for each column}$
\For{$i \gets 0$ to $len\_t-1$}
    \If{$i \neq idx$}
        \For{$j \gets 0$ to $bs-1$}
            \State $all\_t[i][j] \gets \text{random choice of } \{0, num\_timesteps-1\}$
        \EndFor
    \EndIf
\EndFor
\Return $timesteps$
\end{algorithmic}
\end{algorithm}

% TRAINING: Finetune with 300 objects training set, 
% how to build pair: we varies two attributes, roughness and metallicity, from 0 to 1 while keeping the other = 0.
% 1000 epoch, prompt: make it more/less rough/metallic; For example, the roughness of input and ground truth are 0 and 1 for prompt "make it rougher"

% TESTING: we test finetuned model on our validation set. and we set inference steps = 100, text cfg = 7.5, image cfg = 1.0. 
\subsection{Configuration for rendering baseline comparison}
We show comparison against GAN-based material editing \cite{subias2023inthewild}, Null-text inversion with prompt-to-prompt \cite{mokady2022nulltext}, InstructPix2Pix \cite{brooks2023instructpix2pix}, and InstructPix2Pix prompt-only version trained on our data. For rendering baseline comparison, we first performed inverse rendering to acquire the necessary intrinsic attributes and used those to re-render with swapping attributes. By doing this, we ensured our setting was the same as other material editing pipelines. Next, we will go over each of the baselines, and briefly discuss the testing configuration for each of the methods.

\textbf{InstructPix2Pix with our data \cite{brooks2023instructpix2pix}}. 
This method takes an input image and a text prompt for material editing. We compared our method with two versions of InstructPix2Pix: the finetuned version and the original version.
For training, we finetuned the model using a training set of 300 objects and trained for 1000 epochs 
To create pairs for the image editing framework, we varied two attributes—roughness and metallicity—from 0 to 1 while keeping the other attribute at 0. Prompts are built as ``make it more/less rough/metallic.'' 
For example, for the prompt ``make it rougher,'' the roughness of the input and ground truth would be 0 and 1, respectively.
For testing, we evaluated both the finetuned model and the original model on our validation set using the default settings: inference steps set to 100, text CFG set to 7.5, and image CFG set to 1.0.
\begin{figure*}
\begin{center}
\includegraphics[width=0.95\linewidth]{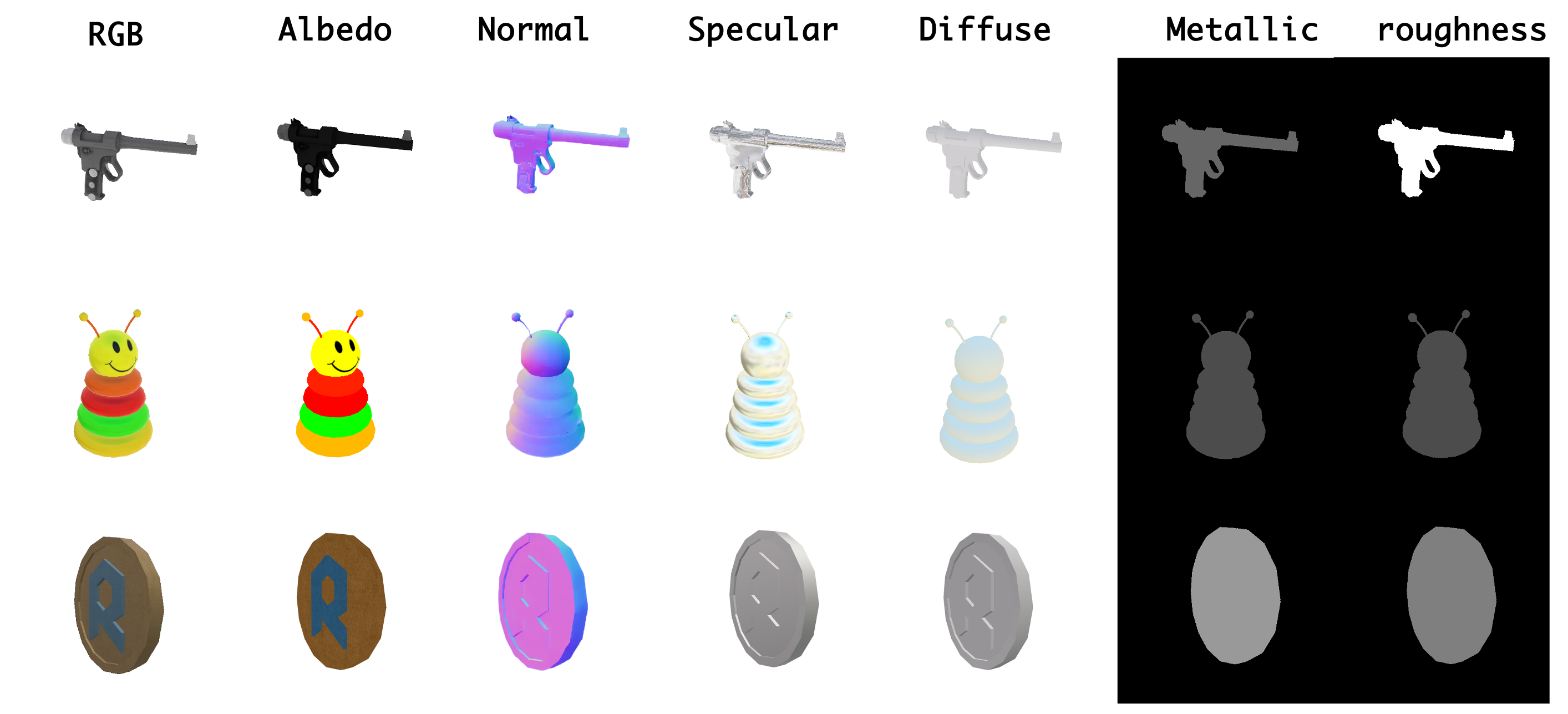}
\end{center}
\vspace{-0.4cm}
   \caption{\textbf{More Qualitative on Inverse Rendering} We included more qualitative cases to demonstrate the ability of Uni-renderer to perform inverse.  \textbf{Best viewed in color.}} 
\label{fig:suppl1}
\end{figure*}

\begin{figure*}
\begin{center}
\includegraphics[width=0.95\linewidth]{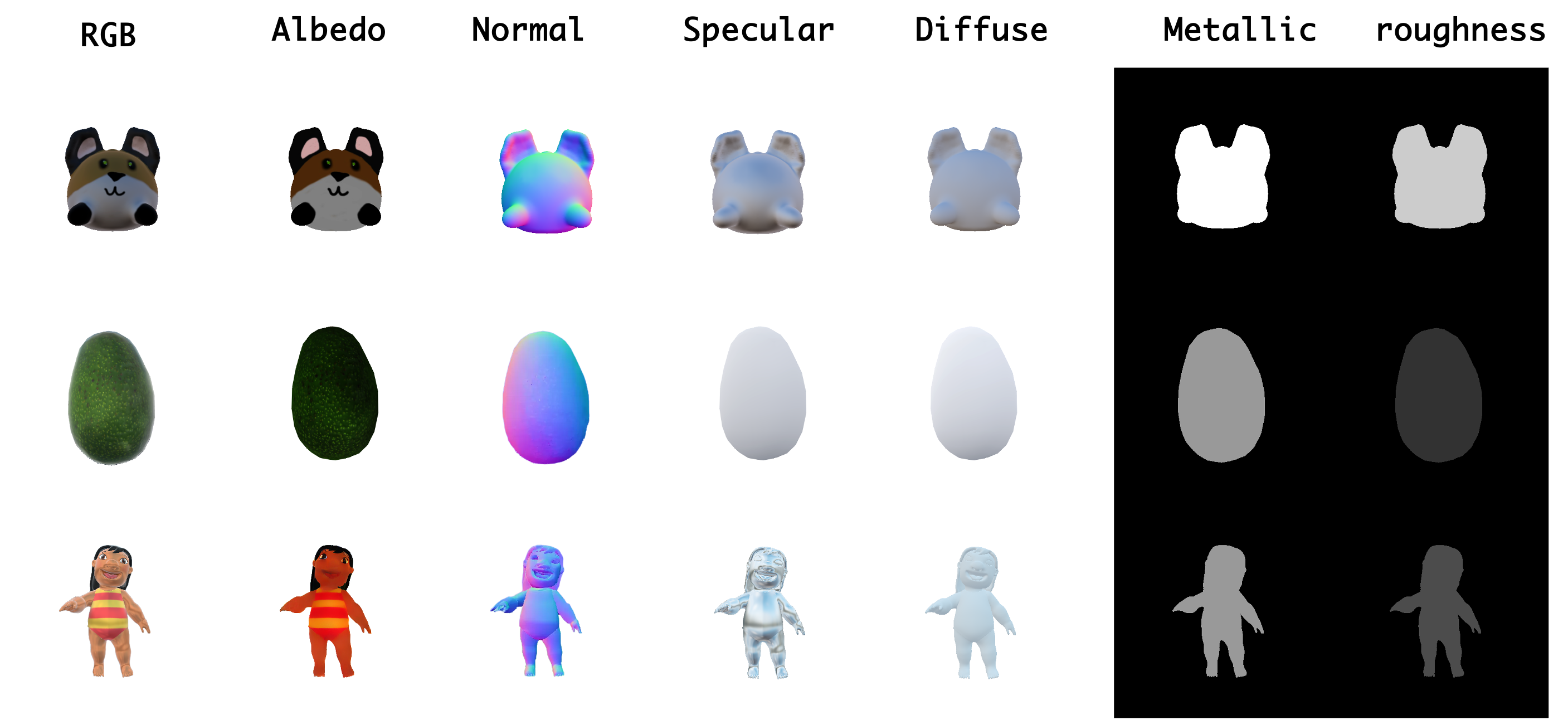}
\end{center}
\vspace{-0.4cm}
   \caption{\textbf{More Qualitative on Inverse Rendering} We included more qualitative cases to demonstrate the ability of Uni-renderer to perform inverse.  \textbf{Best viewed in color.}} 
\label{fig:suppl1}
\end{figure*}

\begin{figure*}
\begin{center}
\includegraphics[width=0.95\linewidth]{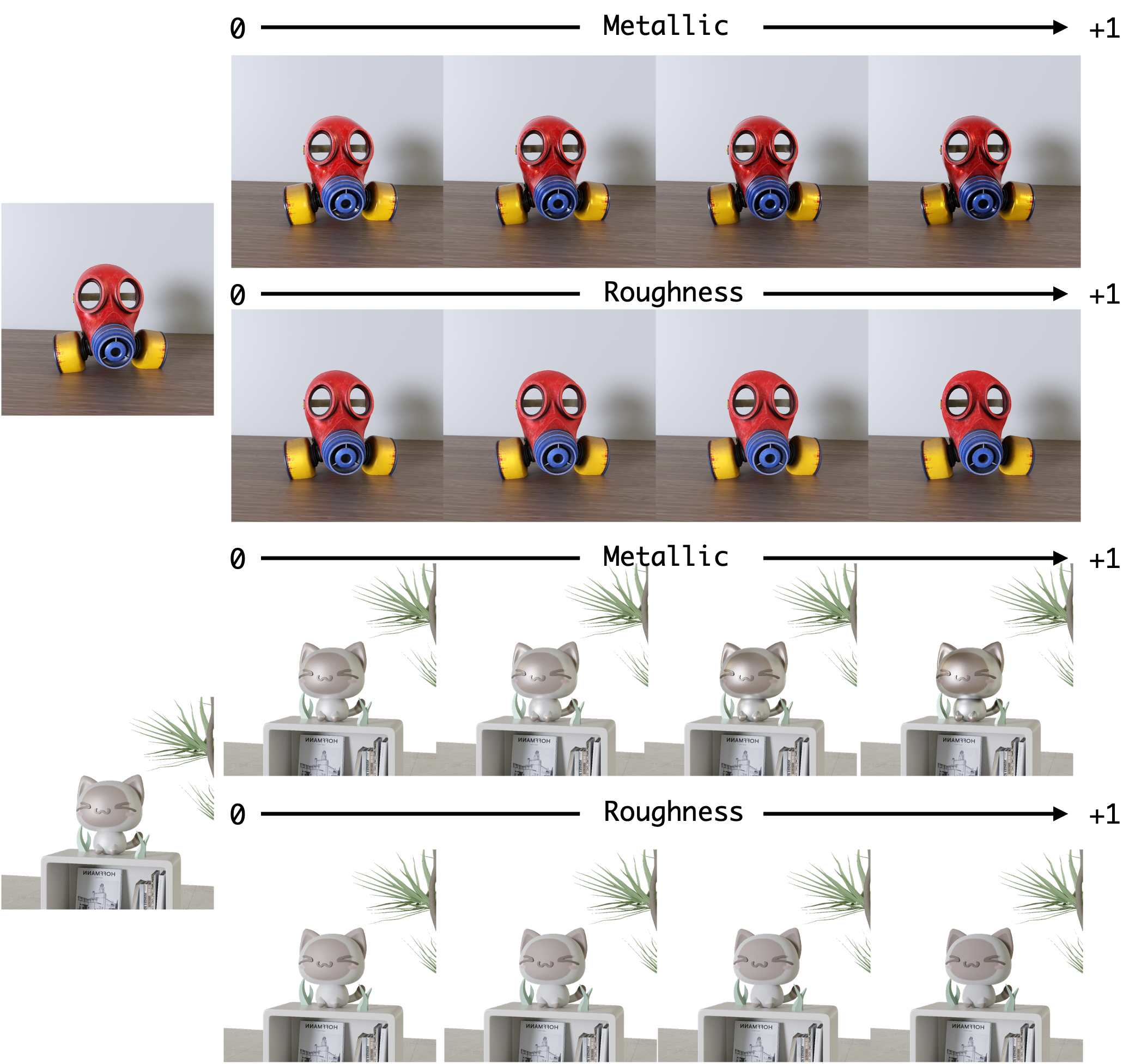}
\end{center}
\vspace{-0.4cm}
   \caption{\textbf{More Qualitative on Accurate attributes editing} We included more qualitative cases to demonstrate the ability of Uni-renderer to perform rendering. The leftmost are reference images, and we have provided both increasing metallic and roughness.   \textbf{Best viewed in color.}} 
\label{fig:suppl1}
\end{figure*}
\textbf{Null text inversion~\cite{mokady2022nulltext}}.
This method, also a prompt-only version, first optimizes the null text embedding to recover the original DDIM latent sequences at inference with a high CFG value. It then performs prompt-to-prompt for image editing.
For optimization, the steps are set to 300, and the prompt pairs used are in the format ``a {object name}'' to ``a metallic/rough {object name}.''
For inference, the steps are set to 50, and the CFG value is set to 7.5.

\textbf{Subias et. al \cite{subias2023inthewild}} The method takes an image as input along with a scalar as input to perform relative material editing for glossiness and metallic. The method requires an input mask for localizing the object in the image. We generated the mask using the provided format input scripts. We input the image with the required transformation and tested it with a scalar of 1.0.

\subsection{More Visual Samples}
In this section, we will include more qualitative samples to further support the effectiveness of our method in terms of inverse and smooth editing in rendering.

% WARNING: do not forget to delete the supplementary pages from your submission 
% \input{sec/X_suppl}

\end{document}